\documentclass[sigconf]{acmart}
 
\usepackage{CJKutf8}
\usepackage{booktabs} 
\usepackage{caption}
\usepackage{subcaption, multirow}
\usepackage{comment}
\usepackage{lipsum} 

\usepackage{arabtex}
\usepackage{utf8}

\usepackage[ruled]{algorithm2e} 

\usepackage{enumitem}
\usepackage{amsmath}
\usepackage{color}

\definecolor{plblue}{rgb}{0.13,0.13,0.7}
\definecolor{pdgrey}{rgb}{0.75, 0.75, 0.75}
\definecolor{pblue}{rgb}{0.13,0.13,1}
\definecolor{pgreen}{rgb}{0,0.5,0}
\definecolor{pred}{rgb}{0.9,0,0}
\definecolor{pgrey}{rgb}{0.46,0.45,0.48}

\usepackage{listings}
\newfloat{lstfloat}{htbp}{lop}
\floatname{lstfloat}{Listing}

\newlist{mylist}{enumerate}{1}
\setlist[mylist]{label*=(RQ\arabic*),ref=RQ\arabic*}
\makeatletter
\newcommand\myitem[1][]{%
  \if\relax\detokenize{#1}\relax
    \item\relax
  \else
    \protected@edef\@currentlabel{RQ#1}%
    \item[(RQ#1)]
  \fi}
\makeatother

\definecolor{lightblue}{rgb}{0.527,0.805,0.977}

\begin{document}
\title[CodeTranslate]{Human Languages in Source Code:\\ Auto-Translation for Localized Instruction}

\author{Chris Piech}
\affiliation{%
  \institution{Stanford University}
}
\email{piech@cs.stanford.edu}

\author{Sami Abu-El-Haija}
\affiliation{%
  \institution{University of Southern California}
}
\email{sami@haija.org}



\begin{abstract}
Computer science education has promised open access around the world,  but access is largely determined by what human language you speak. As younger students learn computer science it is less appropriate to assume that they should learn English beforehand. To that end we present CodeInternational, the first tool to translate code between human languages.  To develop a theory of non-English code, and inform our translation decisions, we conduct a study of public code repositories on GitHub. The study is to the best of our knowledge the first on human-language in code and covers  2.9 million Java repositories. To demonstrate CodeInternational's educational utility, we build an interactive version of the popular English-language Karel reader and translate it into 100 spoken languages. Our translations  have already been used in classrooms around the world, and represent a first step in an important open CS-education problem. 

\end{abstract}

%
%

\maketitle

\section{Introduction}

Reading and writing comments, method names and variable names is a crucial part of software engineering and as such, programs have both a human language, the language of identifiers and comments, in addition to the source-code language (eg Java or Python). This has meant that non-English speakers are often second class citizens when learning to program \cite{jenkins2013english}. In this paper we present a tool for translating a program from one human-language to another to assist in code education, which could reduce the barrier to computer science education for non-English speakers.


The main contributions presented in this paper are:
\begin{enumerate}
    \item Analysis of 1.1M non-English code projects on GitHub
    \item CodeInternational: A tool which can translate code between human languages, powered by Google Translate.
    \item Validation of CodeInternational by evaluating the translation of 1,000 randomly chosen projects from GitHub.
    \item Use of CodeInternational to automatically translate the popular Karel textbook into 100+ languages. We further extend the textbook to parse and run KarelJava code in any language; we report adoption by classrooms around the world.
    
\end{enumerate}


Our human-language code translator was inspired by a desire to make programming more accessible \cite{brinkman2016applying}. An accurate and useful translator would enable faster localization of instruction materials and it would allow learners (as well as practitioners) to translate code that they are working with.

As programming becomes more of a requisite common knowledge skill, we expect coding education to become open-access to everyone. One barrier to this goal is human language. English is currently the modal language of programming instruction perhaps given that the keywords of most of the popular languages, Java, JavaScript etc, are in English (even including Python and Lua, invented in the Netherlands and Brasil respectively). However, a majority of the world, estimated in 2008 at  80\%, can't ``use" English for communication  and substantially more don't speak English as their L1 language (the technical term for one's arterial language, aka, mother tongue)  \cite{crystal2008two}. Should the more than 6 billion non-English speakers learn to program in their native language or in English? This question is debated, which we address in the discussion. 

We take the position that whether or not code instruction is in English, if students do not speak English as their L1 language, their code education would benefit from the ability to translate Code between their preferred language and English.

\subsection{Related Work}
To the best of our knowledge, automatic translation of code between human languages, did not appear in literature, making us hypothesize: it is either difficult, or had remained ignored. Nonetheless, we summarize related work that motivate our contribution.

\textbf{Translation of Text}
automatic translation of natural language has recently achieved high accuracy and is used in highly sensitive contexts \cite{moberly2018doctors, groves2015friend, de2018no}. At the time of writing this article, Google Translate uses Neural Machine Translation  \cite{nmt} to translate pairwise between languages and has become incredibly accurate, at least for languages common on the web \citep{gnmt}. Further research has been done on transliterating text \cite{knight1998machine, arbabi1994algorithms}.
However, current state-of-the-art methods for text translation fail at translating code. Directly running a translation algorithm on code would fail to distinguish between code syntax and identifiers, would not recognize terms embedded in identifiers e.g. with camel case \texttt{getElementAt},
and could produce code with one identifier name having different translations on separate lines.
As such, current automatic text translation, if ran directly on code, would produce malfunctional code. 

\textbf{Code Instruction in Non-English} In 2017, Dasgupta and Hill published seminal work outlining the importance of learning to code in one's own language. They conclude that "novice users who code with their programming language
keywords and environment localized into their home countries'
primary language demonstrate new programming concepts at
a faster rate than users from the same countries whose interface is in English" \cite{dasgupta2017learning}. Since then, there has been a large set of papers expanding on the barriers for non-native English speakers. Guo et al survey over 800 non-English students learning who report on the many challenges that come with not understanding English while coding.
\cite{guo2018non} reinforced by
\cite{dasgupta2018wide, kirkpatrick2011internationalization}.
This has led to preliminary work into translating compiler errors 
\cite{reestman2019native} and advocation for language-free block free programming \cite{banerjee2018empowering}. However, while language-free programming is a great step forward for younger students, it doesn't address the needs of CS1 students who program in common programming languages like Python or Java.
While all of this work motivates our contribution, none has attempted an automatic solution to the problem,
making \textit{crowd-translation} a viable alternative \cite{codeorg_translate}.

\textbf{Mining Github}
To understand the patterns of code that students and practitioners use, we analyze public repositories on GitHub. 
Other researchers also analyzed GitHub, 
sometimes via the dataset and tools provided by
\cite{Gousi13}, 
including work on social diversity of teams \cite{vasilescu2015data} and affiliation influence on code popularity \cite{blincoe2016understanding}. This has led to a set of best practices for navigating the promises and perils of mining GitHub \cite{kalliamvakou2014promises}.
A growing number of students are using GitHub in software
engineering courses \cite{feliciano2016student} which makes it a valuable resource for understanding code of the general population, including students.

\textbf{Code Conversion}
There is a rich literature of work to translate code between \emph{programming languages}, such as C or C++ to Java \cite{terekhov2001automating, vargas2008system}, or even from English to code \cite{little2006translating}. However, the emphasis is often on maintaining efficiency, not on making code readable for students. 
We focus on translating the human language of code. Byckling et al \cite{byckling2005roles} analyze naming conventions of identifiers based their function (fixed, iterators, transformers, etc), and correlate the naming consistency with the students' learning experience. This motivates aspects of our translation. See Section \ref{sec:translating_identifiers}.

\section{Human Languages on GitHub}

\begin{figure}
    \centering
    \includegraphics[width=0.85\columnwidth]{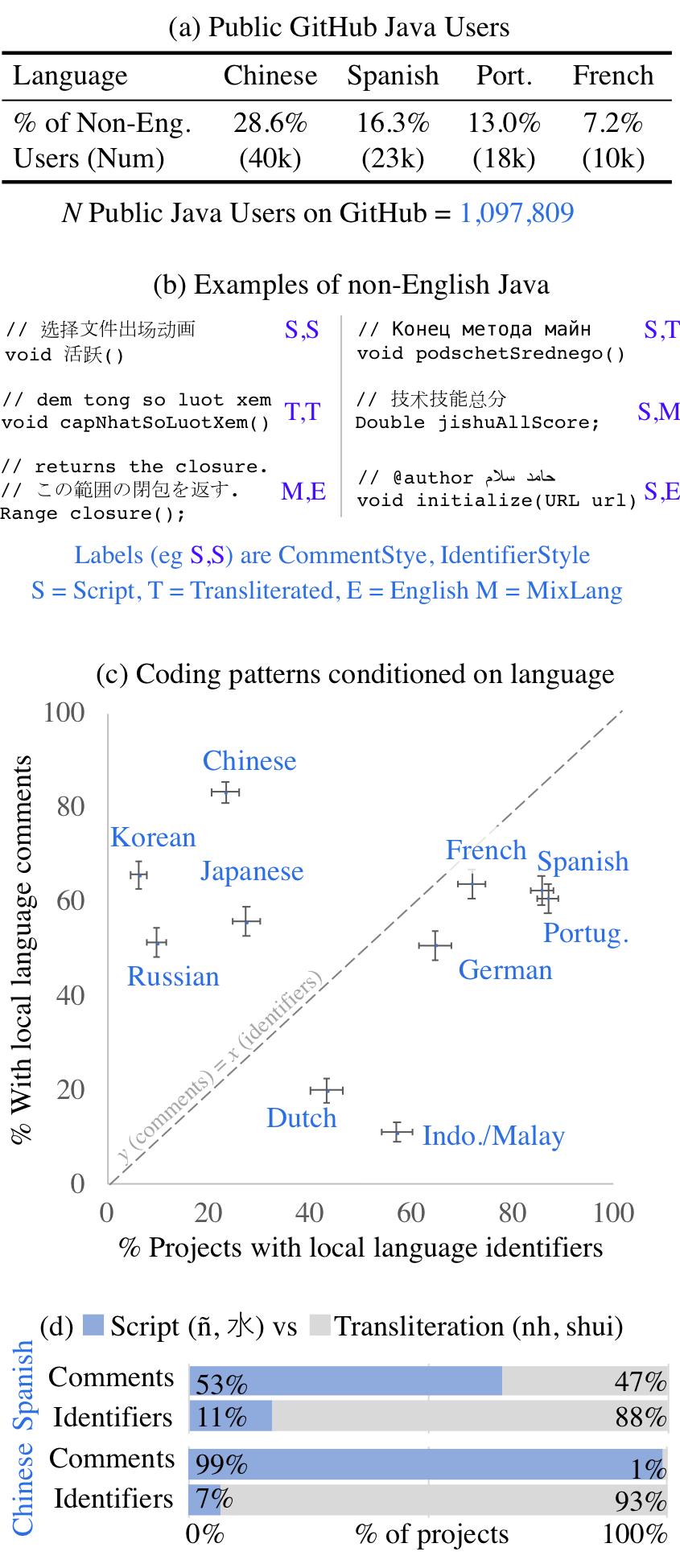}
    \caption{ (a): The four most popular non-Eng languages for Java GitHub commits. (b) Java non-Eng example methods. (c) Use of local language in identifiers and comments conditioned on users speaking different languages. (d) Proportion of non-English projects with script vs transliteration}
    \label{fig:lang-stats}
\end{figure}

How do non-English speakers program in a language like Java, where the keywords and core libraries are written in English?
We employ a data driven approach to tell the story of non-English code and inform the decisions we made in our auto-translator. We analyzed Java repositories on GitHub, the largest host of source code in the world, where 1.1 million unique users host 2.9 million public Java projects. We downloaded and analyzed the human language
used for writing comments (in Java code), naming identifiers (method and variable names), and writing  \textit{git commit} messages.
We focused on Java code as it is both one of the most popular source-code languages on GitHub and in the classroom.  
A selection of results from this study are that:
\begin{enumerate}
\item Non-English code is a large-scale phenomena.
\item Transliteration is common in identifiers for all languages.
\item Languages clusters into three distinct groups based on how speakers use identifiers/comments/transliteration. 
\item Non-latin script users write comments in their L1 script but write identifiers in English.
\item Right-to-left (RTL) language scripts, such as Arabic, have no observed prevalence on GitHub identifiers, implying that existing coders who speak RTL languages have substantial barriers in using their native script in code.
\end{enumerate}
This is, to the best of our knowledge, the first analysis of the human languages on GitHub. See Figure \ref{fig:lang-stats} for an overview.

Users on GitHub do not state their L1 (arterial) language. While a subset of users optionally state their country this is neither common nor reliable. To estimate a user's preferred language we use the language that they use in the git commit message. To find subsets of users who speak a given language, we search for all users who write git commits in that language. We observe that, especially in personal projects, users write commit messages in their L1 language at a higher rate than comments or identifiers. To identify languages we use Google Language Detect which is highly accurate (more so for common internet languages) and can identify languages with non-Roman Alphabet text which has been transliterated, for example it can detect both\begin{CJK*}{UTF8}{gbsn}\hspace{-3mm}算法\end{CJK*} the Chinese characters for ``algorithm" and "suanfa", the Mandarin transliteration, as Chinese\footnote{Google Translate provides a confidence for its language detection. We only consider positive detections with confidence > 0.5. We don't run language detection on ascii strings less than 2 characters long. Identifiers are turned into phrases using  case parsing as described in section 3. All "positive" results are manually verified.}. 

Of the 1.1 million GitHub users, 12.7\% wrote commit messages in non-English languages. Of the non-English languages Chinese was the most common (28.6\% of non-English committers), followed by Spanish, Portuguese, French, and Japanese.  More than 100 languages were detected in commit messages on public Java projects. Figure \ref{fig:lang-stats} contains breakdowns and the appendix contains the full list. This does not match the distribution of non-English in web content (55\% English) with both major and minor languages underrepresented. For example the prevalence of Spanish on GitHub (2.1\%) is about half of webcontent (5.1\% \cite{www-lang-trends}) and further trails native speakers (7.8\% of the worlds population \cite{spanish-speakers}).

Github does not present a random sample of programs written in the world, and we consider the relevant confounds this introduces. 
To that point, we believe the under-representation of certain languages is a form of Survivorship Bias. It suggests that users have found barriers to entry towards joining the GitHub community. Those barriers could derive from the English dominance of programming languages, code instruction, or the github interface.

\subsection{Non-English in Java}

The use of non-English in identifiers and comments is 
large for the population of users who we define as non-English "speakers"
(those who use non-English in their git-commit messages). 90\% of users who use a non-English language in the commit messages also use that language in their comments or as identifiers. We note that, in Java, identifiers can be written in any script. 

Surprisingly, the patterns of non-English usage differs substantially when we condition on users "speaking" different languages. For example, among the detected Spanish speakers, 87.2\% percent of users write identifiers in Spanish. On the other hand, among Chinese users only 23.3\% of users write code with Chinese identifiers (either in Chinese script or ASCII). Figure 1b shows coding patterns conditioned on users speaking different languages. For each language we plot the percent of projects with identifiers in the language against the percent of projects with comments in the language. Languages naturally cluster into three categories: (1) \textbf{Major-Euro-Latin:} languages with high use of non-English identifier including Spanish, German and French (2) \textbf{Non-Latin:} languages in non-latin scripts including Russian and Chinese which have low use of non-English identifiers and (3) \textbf{English-Comment:}  Programmers write their comments in English (> 70\% of projects only have English comments). This group contains many smaller and non-European languages like Dutch and Bahasa Indonesia. \~50\% of projects in this group still uses their L1 language in identifiers.

The use of identifiers in local language (as opposed to English) is very clearly split on whether languages use the Latin alphabet. On average 82\% of projects from users speak languages with different scripts like Chinese, Korean, or Russian have only English identifiers, compared to 12\% of projects from Latin alphabet users ($p < 0.0001$). The percentage of projects with only English comments is roughly correlated to the English Proficiency Index \cite{first2013ef} of the corresponding countries ($\rho = 0.42$ $p < 0.01$).

\subsection{Trans\emph{literation} on GitHub}

Transliteration is the process of transferring a word from the alphabet of one language to another (eg
\includegraphics[height=3mm]{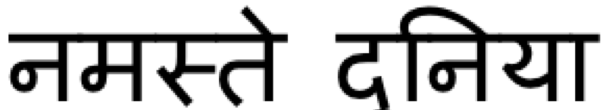} -> namaste duniya).
We observed that most Java code with human languages that have non-ascii scripts like Kanji, Devanagari, or even Spanish accents like ñ, will have been "transliterated" into ascii.

 The Java Language Specification states that, "letters and digits (in identifiers) may be drawn from the entire Unicode character set, which supports most writing scripts". This specification is not widely known, and even if Java supports non-ascii , there can be complexities of file encodings across different operating systems.

We find that regardless of L1 language most users transliterate identifiers: among L1 Chinese speakers, 93\% of projects have identifiers which are only written in ASCII. Similarly in Spanish 88\% of projects have only ASCII identifiers. As a concrete example, in GitHub Java code
"numero" is 3.8x more common than "número". Among comments languages differ greatly: 99\% of Chinese projects have non ASCII comments compared to only 53\% of Spanish. As an example a comment above a method specifies in script that it is calculating the Fibonacci sequence however the method name (an identifier) is transliterated "//\hspace{-3mm}\begin{CJK*}{UTF8}{gbsn}斐波那契\end{CJK*}" however the code uses a transliteration of the phonemes in the script
"public int feibonaqie(int n)". This is a common pattern: Within comments, \hspace{-2mm}\begin{CJK*}{UTF8}{gbsn}计数\end{CJK*} chinese for count), is 4.0x more common  than jishu, the transliteration. However in identifiers jishu is 4.8x more common. The difference in transliteration patterns between Chinese and Spanish suggests a different intent: in Spanish transliteration is used to avoid file encoding errors, in Chinese it is to prevent a mix of scripts among identifiers.

\subsection{Right-to-Left Languages on GitHub}

One question that we did not have a solid pre-conception for was: \textit{How do Java users who speak  languages with right-to-left (RTL) scripts like Arabic, Urdu or Hebrew, write code?}

18,961 users on GitHub report their country as one where a RTL script (Arabic or Hebrew) is the primary script. Those users have 8,060 public Java repositories of which only 50 repositories (0.6\%) have Arabic or Hebrew script (excluding string literals). Of those repositories, only a single Java file had a single identifier written in Arabic and none in Hebrew. It is extremely rare for methods or identifiers to be a mix of  RTL and LTR.

\section{\textsc{Code International}}

The GitHub analysis is coherent with the contemporary narrative: there are perhaps hundreds of millions of learners who will not speak English as their L1 language. For those learners, teachers need a tool to translate code so they can give examples with less congitive load. Similarly students need a tool to understand the non-English code they encounter. Finally, to a growing extent English speakers will begin to interact with code written in other languages.

To adress this need,
we 
designed a tool to help
programmers, regardless of their spoken language, access code in many languages.
The tool, which we call CodeInternational, takes in code written in either Java or Python with comments and identifiers written in a human-language and translates the comments and identifiers into another human-language. It supports the growing set of human languages covered by Google Translate and is adaptive to the particular context of source-code. To translate code, it first parses the code and extracts four types of tokens:
\begin{itemize}[leftmargin=3mm]

\item \textbf{Comments}: inline or multi-line comments. Their purpose is for the programmer to communicate to programmers (including herself) on the purpose of code sections. 

\item \textbf{Immutable}: consisting of language  keywords (\texttt{while}, \texttt{void}, etc), and identifiers imported from libraries that are external to the code being translated (e.g. \texttt{FileReader} of \texttt{java.io}). By default this group is not translated.

\item \textbf{Target identifiers}: including variable and function names that are defined in the code base undergoing translation. 
\item \textbf{String literals}: In some cases a user may want String literals to be translated, other times they should be unchanged.  
\end{itemize}

\begin{figure}
    \centering
    \includegraphics[width=\columnwidth]{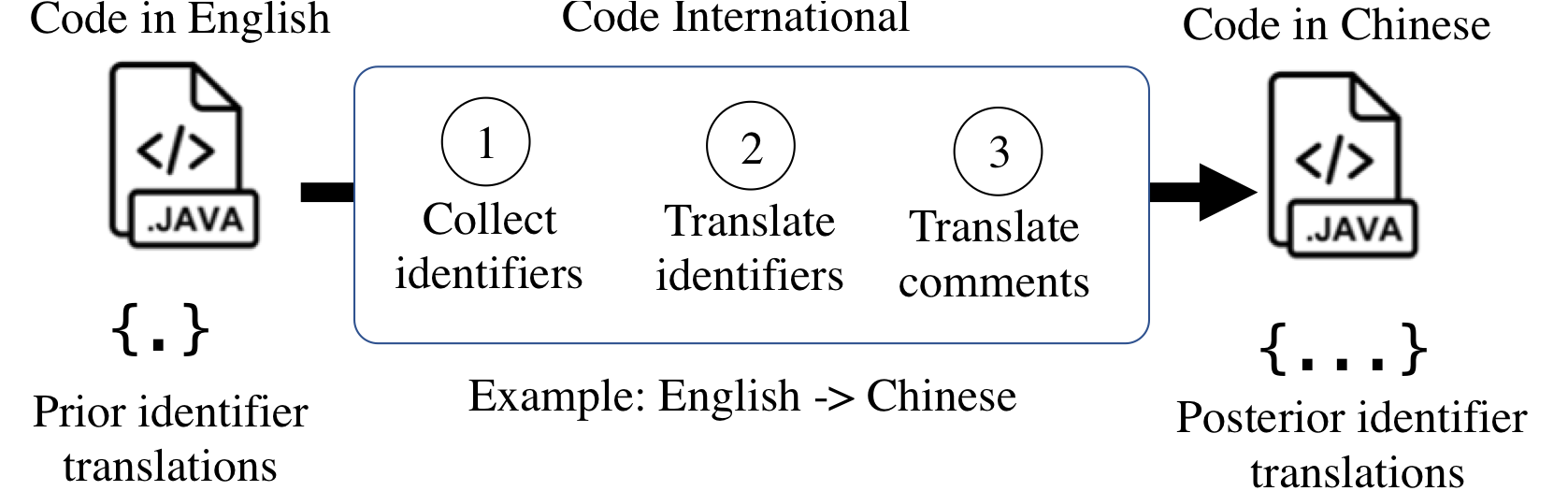}
    \caption{High-level of how CodeInternational work}
    \label{fig:api}
    \vspace{-4mm}
\end{figure}

\begin{figure*}
    \centering
    \includegraphics[width=0.95\linewidth]{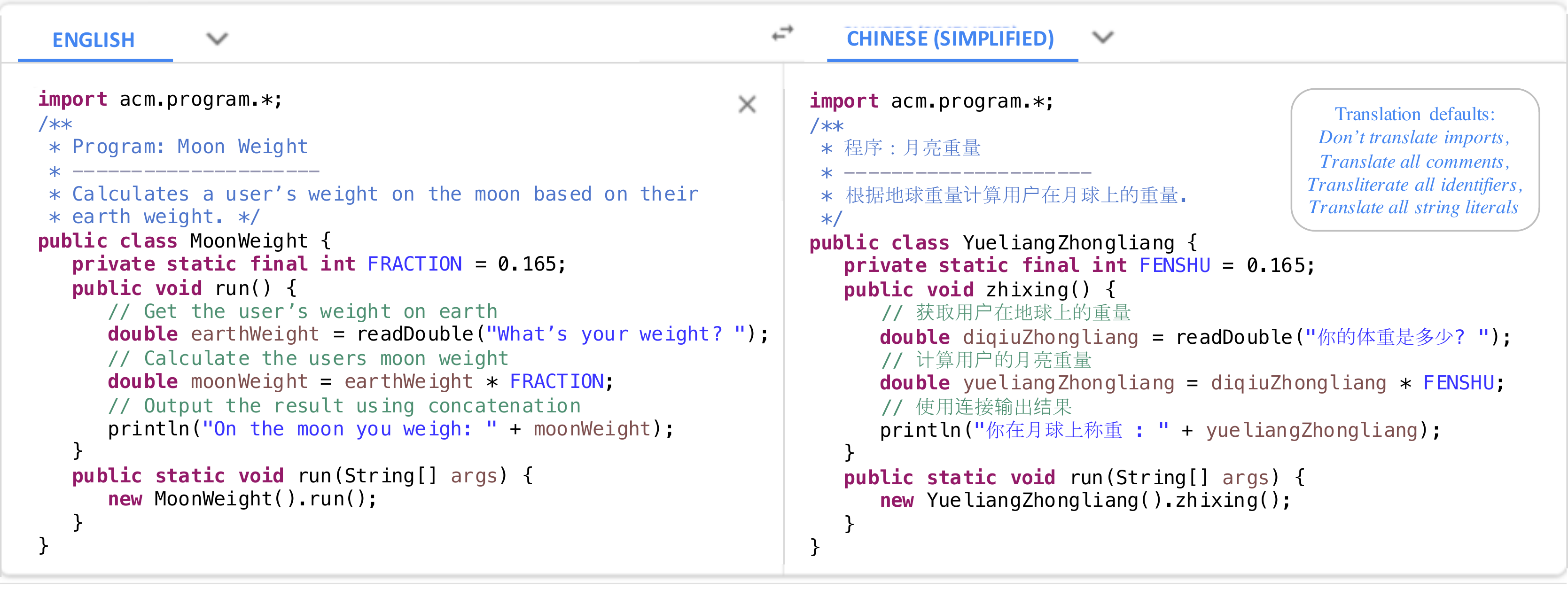}
    \vspace{-5mm}
    \caption{An example of using CodeInternational to translate a simple Java program from English to Chinese.}
    \label{fig:exampleTranslation}
    \vspace{-3mm}
\end{figure*}

Our translation algorithm is as follows. We (1)  collect all of the target identifiers defined in the codebase and (2) translate them (enforcing that if two identifiers have the same name, they are given the same translation). Once the identifiers are translated we (3) translate the comments preserving structure and references to identifiers. (4) Finally string literals are optionally translated. See Figure \ref{fig:api} for a highlevel depiction and Figure \ref{fig:exampleTranslation} for a concrete example. Each of these steps has surprising challenges. In this section we cover the corresponding solutions we developed. The mapping of identifier translations that the tool decides on is preserved to assist any external source which needs to refer to the newly translated identifiers (such as text in a text-book or code in a related project).

CodeInternational is implemented in Python. Tokenization is performed using a modified version of "Javalang" (for Java) and the "Parser" library (for Python). Supporting other programming languages  requires a small amount of extra work\footnote{We expect C, C++ and Javascript to be ready by the camera-ready deadline.}.

\subsection{Translating Identifiers}
\label{sec:translating_identifiers}
In order to properly translate identifiers, we consider the following:

\vspace{2mm}
\noindent
\textbf{Identifier segmentation:} Translating an identifier using a tool like Google Translate does not work by default as identifiers are often composed of unsegmented words. For example: getFavoriteNumber is readable to a human as "get favorite number" but is not parsable by an online translator. 
We segment identifiers using naming conventions  (e.g. \texttt{camelCaseVariable}, \texttt{PascalCaseClass}, \texttt{UPPERCASE\_CONSTANT}).
We thus segment identifiers into phrases which we feed into an automatic translator. We then recombine the translated phrase using the original casing convention. For example, to translate the method name identifier "turnAround" into Spanish: "turnAround" is segmented into "turn around" which is translated into "media vuelta" which is formatted into the original camelCase "mediaVuelta". Advances in artificial intelligence for word segmentation enable a future version of this tool to break up words \emph{without} a given segmentation (eg "turnaround").


\vspace{2mm}
\noindent
\textbf{Verb prior:}
The correct translation for a phrase can be ambiguous, especially without context. As an example the method "move" translated into Spanish could be translated into a noun ("movimiento", movement) or a verb ("moverse"). For method identifiers there is an implicit context that an action is being performed. We incorporate this context by placing a prior on the first word being a verb. Thus, for example, when we translate "move()" into Spanish we chose "moverse()" instead of "movimiento()", the noun movement, as Google suggests. 

In addition to knowing the translations of methods should start with verbs, we also have a select number of reasonable tenses for the verb: infinitive (eg "toMove"), third person present (eg "moves" as in "he moves") and imperative (eg "move"). In most languages, including English, we translate verbs with a prior that they be the imperative tense. In English you would expect a method to be "getObject()" the imperative. However some languages, especially Romance languages, use the infinitive of the verb: as an example, "obtener" the infinitive of "obtain" is 200x more common on GitHub then "obtenga" the imperative. 

\vspace{2mm}
\noindent
\textbf{Translating short identifiers:}
Short variable names that are used for mathematical symbols or as iterators should not be translated. This is especially important to pay attention to for the cannonical for loop identifier "i". For example translating the code "for(int i = 0; i < 10; i++)" into Spanish should not produce "for(int yo = 0; yo < 10; yo++)" even though "yo" is the translation of the pronoun "I". We only translate identifiers which are at least two characters long. This exception has its own edge-case: CJK (Chinese, Japanese Korean) identifiers can be  non-mathematical names even if only a character long.

\subsection{Translating comments}
Once we have finished translating identifiers, we translate the comments in a program. Translating comments has two complexities: (1) we would like to maintain the comment structure, eg if it is a block javadoc comment, we would like to reserve the column of '*'s on the left margin of the comment and (2) we want references to identifiers to be translated exactly as they were in the code.

To translate a comment we classify the structure (eg JavaDoc, BlockComment PythonDocString). We then strip the text out, translate it, and reformat it back into the same structure. For multi-line comments we are conscious not to increase the maximum length of a line, taking into account the wider width of CJK characters.

\subsection{Translating Right-to-Left languages}
Arabic, Hebrew, Farsi, and Urdu are popular right-to-left (RTL) natural languages.
When translating code to RTL languages, comment can be translated (mixing RTL within the left-to-right syntax) and optionally transliterated (keeping left-to-right flow). Some of the difficulty in RTL transliteration is in distinguishing between short- and long-vowels. Further, these languages contains consonant that cannot be described using Latin alphabets, which are generally represented with numbers in the transliteration culture -- e.g. 7 for
\setcode{utf8}
\< ح >
, which is closest to Latin alphabet ``h'' e.g. in ``A\textbf{h}mad''.

When translating non-Latin scripts which are LTR we give the user the option to transliterate identifiers and separately, to transliterate comments or not. Transliteration is currently supported in Arabic, Chinese, Hebrew, Japanese, Korean, and Russian.

\subsection{Prior and posterior translations}
Translations of code need to be coherent with respect to other translations of written text (or other files) that refers to the code. To that end our translator takes in, and uses, a preset identifier translation map and returns the translations it made. This system enables having humans override translations, translating text-books with text that references embedded code and more.



\section{Translating Github}

\begin{figure*}
    \centering
    \includegraphics[width=0.95\linewidth]{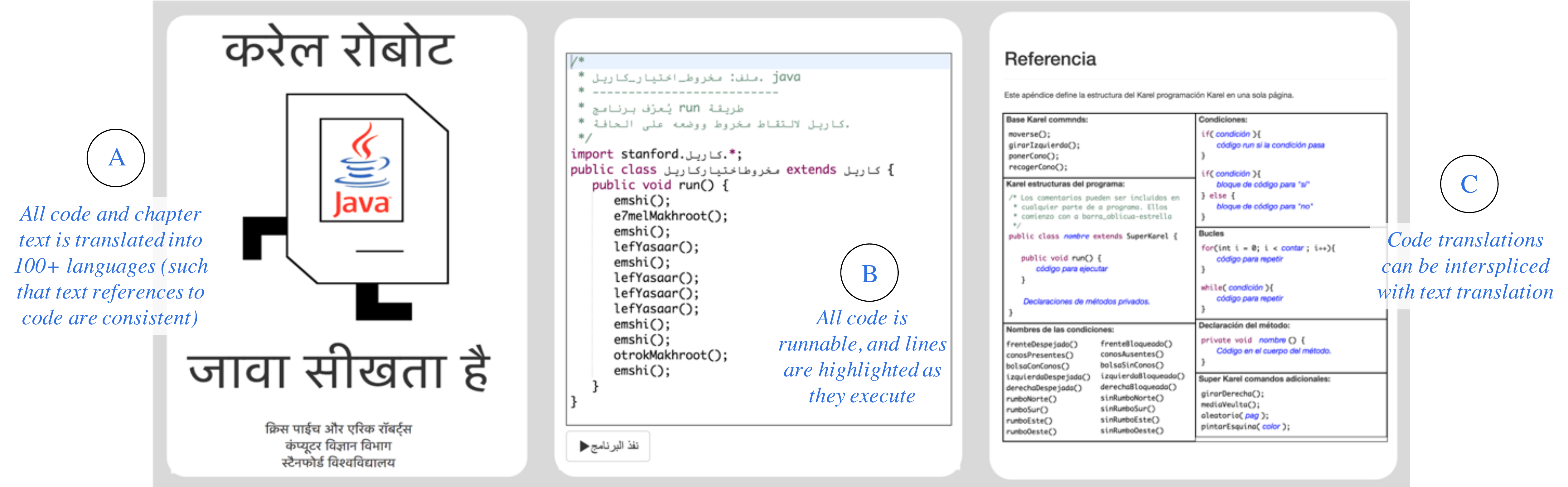}
    \vspace{-3mm}
    \caption{
       Three screenshots from the Karel eReader, translated both into 100+ languages for Java and Python:
       Left: intro page in Hindi; Middle: code translated into Arabic with transliterated identifiers; Right: reference in Spanish.
        }
    \label{fig:karel}
    \vspace{-3mm}
\end{figure*}

How good is a translation of source-code from one human language to another? Evaluating quality of a translation is hard without a large collection of native speakers and since we are powered by Google, evaluation can devolve into evaluating how accurate Google Translation is. Such an evaluation is a moving target: Google Translation is perennially improving.

To evaluate out translator we randomly selected 1,000 (1k) single file projects from public GitHub Java and translated them into the languages: Chinese, Spanish and Arabic. We measure (1) how often the translated code still compiles and (2) what percent of identifiers that we attempt to translate are translatable.

Of the 1k projects 100\% maintained their ability to be compiled regardless of whether we translated or transliterated the comments or identifiers. From the 1k projects 91\% of the identifiers were able to be translated. The nine percent that were not able to be translated were mainly abbreviations (such as users who named a variable frac instead of fraction or pct instead of percent). This is an opportunity for future work. Overall the results paint the picture of a functioning tool which is ready for use.

\section{International Karel}
Our motivation for developing an automatic human-language code translation tool was to support education for non-English speakers. To that end we used CodeInternational to translate a web-version of the popular Karel the Robot learns Java reader by Eric Robers \cite{roberts2005karel} a textbook for a Karel the Robot, a grid world robot invented by Richard Pattis \cite{pattis1995karel} to help CS1 students learn to program. Karel has been the inspiration for assignments on platforms such as Code.org and CodeHS and is a staple of the first weeks of CS1 \cite{becker2001teaching}. 

We translated a Karel reader in Python and Java to 100 languages. The translated web-reader is free to use, and is hosted at [redacted]. At time of publication the reader has been public (without advertising) and has already been used by over 3,000 people from 50 countries. 
With permission from Eric Roberts, we first made an eBook version of his Karel reader and simplified the English used \cite{ericson2016identifying}.  The reader merges text and code in a seemless fashion. Then, for each language: we (a) translated code on each chapter using CodeInternational and (b) translated the reader text such that any reference to identifiers in the example code would use the same translations. In order to have text which is consistent with the corresponding code we heavily rely on the "Posterior identifier translation map" from CodeInternational's translations.


\subsection{Line-highlighting in any language}

To make the Karel reader a fantastic learning experience we made it so that each code-snippet is runnable. When run, the program \textbf{\emph{executes the code and highlights the corresponding lines}} as the program is run, regardless of the complexity of the program's control flow. In order to line-highlight we parse and compile the Python-Karel or Java-Karel programs using an engine written in JavaScript. Our line-highlighter builds upon the compiler
described in \textit{Informatics Education using Nothing but a Browser} \cite{piech2011informatics}.

Our Karel reader can run and line-highlight in any human-language that we translate into. For example our compiler can execute and line-highlight the command "moverse()" if the code is written in Spanish, "\begin{CJK*}{UTF8}{gbsn}\hspace{-3mm}移动\end{CJK*}()" if the program is written in Chinese, "emshi()" if the program is written in Arabic, or "move()" if the Karel program is written in English. We chose to only transliterate commands for RTL scripts. Figure \ref{fig:karel} shows three screenshots from the international Karel reader, though of course a PDF is unable to capture the ability of the reader to line-highlight code.

\subsection{Usage in Classrooms}
We know of four classes where the internationalized Karel eReader has been used. These classes are around the world in: Istanbul, Bogota, Prague and [Redacted]. The eReader has been visited by >1k users in 3 months and both the English and the non-English version of the website have a high average session duration (9.7 min and 10.1 min respectively). Moreover, the tool has been used to translate the CSBridge curriculum website and assignments; HTML that mixes code and description (used by 400 students / year).


\section{Discussion}

Whether English should be used as the sole language of instruction has been debated.
\textbf{Case for code instruction in English}: In order to program professionally, one will have to interact with keywords and libraries that are written for English speakers.
English is \textit{the}  language of code, and it is practically required from anyone who wants to interact globally: correspond via email, read stack-overflow, watch educational videos, travel, etc. For classrooms where English is the main form of instruction, but students are not yet fluent, CodeInternational can be used to assist learning English and learning to program. Students could improve their English through coding, e.g. by placing English code against their L1 code, side-by-side.
\textbf{Case for instruction on transl(iter)ated code}: 
On the other hand, people argue that it is beneficial for students  to have much of their coding instruction in their L1 language, and doing so benefits access to CS.
The primary reason for this intuitive: the cognitive-load of learning to program is already high. Moreover, if students learn coding using their L1 language and enjoy it, they become intrinsically motivated to learn English, knowing that English would broaden their access to learning material (learning earning a language, with no short-term motives, could be dull  especially for young students). In this context CodeInternational can help students who are interacting with libraries in English. Perhaps more importantly our tool can help teachers rapidly develop localized content that builds off English content. The alternative: manual-translation of API, code-examples and website text, can be a huge barrier to translating material. Finally, our tool builds off GoogleTranslate, which is high accurate, but charges \$1 per 50,000 characters. A free version would have a huge impact on utility.

We call for \textbf{future work} from tool experts, for extending popular code-editors (e.g. vim, XCode, Visual Studio, Eclipse) to integrate with our APIs for back-and-forth translation and side-by-side display. Optionally, integrating with automatic text-to-speech (e.g. \citep{wavenet-text2speech}) could allow students learn English pronunciation of code components. Moreover, one remaining feature in automatic human-translation of code is identifier consistency: if two identifiers have specific terms in common, eg getHeight, setHeight, we would like the translation of "height" to be consistent. While they are often consistent in our work, it is not enforced. Full consistency is hard, but not impossible, with modern neural machine translation.




\section{Conclusion}

We analyze millions of non-English Java programs on GitHub to inform our understanding of patterns of human-language and make some surprising observations. We build CodeInternational, an open-source tool which can translate Java or Python code between human languages. We evaluate our tool and use it to make an internationalized Karel eReader (with runnable code) in 100+ languages. Our tool  is already being used in classrooms around the world, a trend we hope to continue supporting.

\subsection*{Acknowledgements}
We would like to graciously thank XXX and YYY for contributing code to this translation project. We would also like to thank ZZZ for her inspiration. We also thank the WWW teachers for educating students around the world in their local language.

\bibliographystyle{ACM-Reference-Format}
\bibliography{biblio}





\end{document}